# Bayesian Classification and Feature Selection from Finite Data Sets

**Frans M. Coetzee**
NEC Research Institute
4 Independence Way
Princeton, NJ 08540

**Steve Lawrence**
NEC Research Institute
4 Independence Way
Princeton, NJ 08540

**C. Lee Giles**
NEC Research Institute
4 Independence Way
Princeton, NJ 08540

## Abstract

Feature selection aims to select the smallest subset of features for a specified level of performance. The optimal achievable classification performance on a feature subset is summarized by its Receiver Operating Curve (ROC). When infinite data is available, the Neyman-Pearson (NP) design procedure provides the most efficient way of obtaining this curve. In practice the design procedure is applied to density estimates from finite data sets. We perform a detailed statistical analysis of the resulting error propagation on finite alphabets. We show that the estimated performance curve (EPC) produced by the design procedure is arbitrarily accurate given sufficient data, independent of the size of the feature set. However, the underlying likelihood ranking procedure is highly sensitive to errors that reduces the probability that the EPC is in fact the ROC. In the worst case, guaranteeing that the EPC is equal to the ROC may require data sizes exponential in the size of the feature set. These results imply that in theory the NP design approach may only be valid for characterizing relatively small feature subsets, even when the performance of any given classifier can be estimated very accurately. We discuss the practical limitations for on-line methods that ensures that the NP procedure operates in a statistically valid region.

## 1  INTRODUCTION

Intense interest in data mining and web searching has renewed interest in the problem of feature selection. In feature selection, the objective is to select the smallest subset of features that meet classification performance requirements, thereby reducing computational complexity and ensuring generalization [1–3].

Feature selection methods summarize the achievable classification performance on a possible subset of features, and use this information to guide the search through the feature power set. In contrast to most analyses, we consider the most general problem where performance summarization depends only on the statistical structure of the data, not the classifier architecture (as occurs in wrapper approaches), a specific operating point cannot be assumed, and data is assumed finite. Such problems usually arise when features are selected in advance of complete knowledge of the application, or when data will be collected for use in a range of applications and risk functions, an example being data selection for futures risk assessment.

Theoretically, the minimal representation of classification performance on a feature subset meeting the above requirements is the Receiver Operating Curve (ROC) [4,5]. The ROC curve is important since it summarizes in a one-dimensional function the optimal classification performance achievable by all possible classifiers, independent of the dimension of the feature space. Therefore, all cost functions can be optimized when this curve is known. When the class conditional statistics are known, the Neyman-Pearson[1] (NP) design procedure found in any classification textbook is the standard way for calculating the ROC.

In practice, ROC calculations are performed by applying the NP design procedure to density estimates obtained on finite data sets. The performance estimates obtained for each feature subset are subject to error. As a result, the user might incorrectly rank different feature subsets, causing the search procedure to fail and yielding substantially inferior feature subsets and classifiers.

In this paper we derive quantitative bounds relating the size of the subset, the amount of data and the confidence in the performance estimates of the ROC curve that can be obtained using NP design. We show that estimates of performance for a classifier do not depend on the size of the feature set. However, the NP design procedure itself is subject to severe error as the size of the feature sets increase; therefore, its efficiency at removing suboptimal classifiers from consideration is impaired. As a result, the user might incorrectly conclude that the performance differences between different feature subsets is statistically insignificant simply because of an inability to find the optimal set of classifiers. Our analysis points to methods for making the NP procedure more robust, by quantifying the confidence in the NP design procedure on-line.

---

[1] The NP procedure is often inaccurately simply called Bayesian design.

Our work complements previous work [1, 6] that analyzed estimation errors in the search across feature subsets. These researchers noted that finding the optimal classifier on a feature subset imposes limitations on feature selection; our analysis quantifies this problem for probably the most fundamental design approach. We also show that this difficulty in finding the optimal classifier on a subset may fundamentally limit the size of the feature subsets that can be absolutely ranked.

The outline of the paper is as follows. In Section 2 we formalize and analyze the NP design procedure given the class probabilities (or infinite data). Section 3 addresses the problem of NP design on finite data sets. Section 4 discusses the propagation of error in typical problems. Section 5 describes data-adaptive methods for performing feature selection in a statistically valid way.

## 2 ROCs FROM INFINITE DATA

This section considers two class classification when the class conditional densities are known to arbitrary precision, as might be the case when densities were obtained from infinite cooperating data sets. While the theory is well known, the formalization of the Bayesian design process as a search procedure is somewhat uncommon. However, this perspective clarifies the error propagation in the procedure.

### 2.1 BAYESIAN DESIGN AS SEARCH

Assume that we are given $N$ possible features, $Q^* = \{x_0^*, x_1^*, \ldots x_{N-1}^*\}$. For simplicity, we assume that the features are binary, that is $x_i^* \in \{0, 1\}$. The analysis carries over in a straightforward manner when $x_i^*$ assume values in any finite alphabet. We assume two hypotheses $H_0$ (false class) and $H_1$ (true class) on the input space $\chi(Q^*) = \prod_{j=0}^{N-1}\{0, 1\}$ with class conditional probabilities $\mathcal{P}\{x^* \mid H_0\}$ and $\mathcal{P}\{x^* \mid H_1\}$ respectively. For a given subset $Q \subseteq Q^*$ consisting of $l$ features, we obtain a set of possible inputs $\chi(Q) = \{x = (x_0, x_1, \ldots x_{l-1}) | x_i \in Q^*\}$. Each sample $x \in \chi(Q)$ can be denoted by a bit string of length $l$. We adopt the convention of denoting vectors using symbols without subscripts, with vector elements indicated by subscripts. We further frequently associate the bit string $x$ with its integer mapping, e.g. $x = (x_0, x_1) = (1, 0) = 2$.

Given a feature subset $Q$, a classifier function $\Gamma : \chi(Q) \to \{0, 1\}$ assigns labels, either 0 or 1, to every element in the binary sequence space $\chi(Q)$, thereby forming decision regions $\mathcal{L}(\Gamma)_0$ and $\mathcal{L}(\Gamma)_1$ in $\chi(Q)$ for the two classes respectively. The set of all inputs $x \in \chi(Q)$ consists of $2^l$ possible bit strings. There are therefore $2^{2^l}$ distinct different classification rules $\Gamma_j, j = 0, 1, \ldots 2^{2^l} - 1$ for separating the two classes.

The classifiers $\Gamma_j$ can be ordered so that the labeling induced by $\Gamma_j$ on $\chi(Q)$ is the binary expansion of the integer $j$. Each of these decision rules yield a probability of false alarm $P_f(\Gamma)$ and of detection $P_d(\Gamma)$, defined by

$$P_f(\Gamma) = \sum_{x \in \mathcal{L}(\Gamma)_1} \mathcal{P}\{x \mid H_0\} \quad (1)$$

$$P_d(\Gamma) = \sum_{x \in \mathcal{L}(\Gamma)_1} \mathcal{P}\{x \mid H_1\} \quad (2)$$

The set $AOS = \{(P_f(\Gamma_j), P_d(\Gamma_j))\}$ of the operating points defined by the $2^{2^l}$ binary mappings on a feature set is referred to as the Achievable Operating Set. By switching between the outputs of two classifiers with some fixed probability, any operating point $(P_f, P_d)$ on the line connecting the operating points of the two classifiers can be produced [5]. Hence, any operating point within the convex hull of the AOS can be obtained. For each feature set, the function $ROC = \{(P_f, P_d)\}$ where $P_f$ is the false alarm rate and $P_d$ is the *maximal* probability of detection achievable *over all possible sampled combinations of functions* $\Gamma_j$ at that value of $P_f$ is called the Receiver Operating Curve (ROC). The ROC efficiently summarizes the inherent difficulty in separating the two classes on a given subset. The subset of the $2^{2^l}$ classifiers $\Gamma_j$ lying on the ROC will be referred to as the ROC support classifiers.

Naively searching for the ROC requires finding all $2^{2^l}$ operating points and calculating the convex hull. This approach is infeasible due to the rapid growth in the value of $2^{2^l}$ as $l$ increases. For example, when $l = 4$, $2^{2^l} = 65536$, while when $l = 5$, there are $2^{2^l} \simeq 4.3 \times 10^9$ functions to consider. The Neyman-Pearson (NP) design procedure provides a solution to the problem of efficiently obtaining the ROC. According to this theory the likelihood ratio function $\zeta : \chi \to \Re^+$

$$\zeta(x) = \mathcal{P}\{x \mid H_1\}/\mathcal{P}\{x \mid H_0\} \quad (3)$$

transforms the bit string $x$ into a scalar random variable for which the decision regions are contiguous and are separable by a single threshold on $\zeta$, as determined by the false alarm rate. Since the feature space $\chi$ is countable, there are at most $2^l$ unique decision region allocations maximizing the probability of detection for the associated level of false alarm. In particular, each of these optimal classifiers (denoted $\Gamma^m$) is associated with one of the thresholds in the finite set

$$Z = \{\zeta | \zeta = \mathcal{P}\{x \mid H_1\}/\mathcal{P}\{x \mid H_0\} \; x \in \chi\} \quad (4)$$

The values $\zeta(x)$ obtained from $p(x|H_1)$ and $p(x|H_0)$ on the elements of $\chi(Q)$ are sorted and a function $\alpha$ is defined that associates with each value of the alphabet the rank of the threshold (in decreasing order),

$$\alpha(x) = \mathcal{I}\{\zeta(x)\} \quad (5)$$

where $\mathcal{I}$ denotes the index of the element after sorting. The ROC curve is then traced by the set of assignments

$$\Gamma^m(x) = \begin{cases} 1 & \alpha(x) < m \\ 0 & \text{else} \end{cases} \quad (6)$$

for $m = 0, 1, \ldots 2^l$. In other words, the ROC support classifiers are produced by successively changing the labeling of an additional bit in the classifier integer expansion in the order specified by the likelihood ratio ranking.

To make the above abstraction more intuitive we present a brief example for characterizing a simple two-feature test subset. The true density functions are

| $x$ | 0 | 1 | 2 | 3 |
|---|---|---|---|---|
| $p(x\|H_1)$ | 0.30 | 0.35 | 0.20 | 0.15 |
| $p(x\|H_0)$ | 0.15 | 0.25 | 0.40 | 0.20 |

The complete list of all $2^{2^2}$ decision region assignments for this subset, and the associated $P_f$ and $P_d$ values; are

| $\Gamma(0)$ | $\Gamma(1)$ | $\Gamma(2)$ | $\Gamma(3)$ | $P_f$ | $P_d$ |
|---|---|---|---|---|---|
| 0 | 0 | 0 | 0 | 0.00 | 0.00 |
| 1 | 0 | 0 | 0 | 0.15 | 0.30 |
| 0 | 1 | 0 | 0 | 0.25 | 0.35 |
| 1 | 1 | 0 | 0 | 0.40 | 0.65 |
| 0 | 0 | 1 | 0 | 0.40 | 0.20 |
| 1 | 0 | 1 | 0 | 0.55 | 0.50 |
| 0 | 1 | 1 | 0 | 0.65 | 0.55 |
| 1 | 1 | 1 | 0 | 0.80 | 0.85 |
| 0 | 0 | 0 | 1 | 0.20 | 0.15 |
| 1 | 0 | 0 | 1 | 0.35 | 0.45 |
| 0 | 1 | 0 | 1 | 0.45 | 0.50 |
| 1 | 1 | 0 | 1 | 0.60 | 0.80 |
| 0 | 0 | 1 | 1 | 0.60 | 0.35 |
| 1 | 0 | 1 | 1 | 0.75 | 0.65 |
| 0 | 1 | 1 | 1 | 0.85 | 0.70 |
| 1 | 1 | 1 | 1 | 1.00 | 1.00 |

These 16 operating points (AOS) are also shown in Figure 1. The ROC classifiers are produced sequentially by changing one bit in the binary expansion based on the likelihood sort $\alpha$:

| $x$ | 0 | 1 | 2 | 3 | $P_f$ | $P_d$ |
|---|---|---|---|---|---|---|
| $\zeta(x)$ | 2.00 | 1.40 | 0.50 | 0.75 | | |
| $\alpha(x)$ | 0 | 1 | 3 | 2 | | |
| $\Gamma^0(x)$ | 0 | 0 | 0 | 0 | 0.00 | 0.00 |
| $\Gamma^1(x)$ | 1 | 0 | 0 | 0 | 0.15 | 0.30 |
| $\Gamma^2(x)$ | 1 | 1 | 0 | 0 | 0.40 | 0.65 |
| $\Gamma^3(x)$ | 1 | 1 | 0 | 1 | 0.60 | 0.80 |
| $\Gamma^4(x)$ | 1 | 1 | 1 | 1 | 1.00 | 1.00 |

Neyman-Pearson design can therefore be viewed as a search procedure whereby the classifier functions $\Gamma^m$ that support the ROC curve can be obtained sequentially in order of increasing probability of false alarm. The problem of finding the function that maximizes the $P_d$ at a given value of $P_f$ is reduced from searching a space of dimension $2^{2^l}$ to one of searching a space of dimension $2^l$, an enormous reduction in complexity.

## 2.2 MAXIMAL ERRORS IN ROC CURVE

There is a one-to-one relationship between the ranking $\alpha$ and the set of classifiers $\{\Gamma^m\}$ that will be produced. When two bins are interchanged in the ranking procedure, the true performance achieved is inferior to the ROC operating curve. This error exists over the range of false alarm corresponding to the two classifiers where neither bin is selected, and where both bins are selected. The true operating curve moves down in detection rate and up in false alarm rate. Assume that for two bins $x_a$ and $x_b$, there are real numbers $\eta_1$ and $\eta_0$ such that $\mathcal{P}\{x_a \mid H_1\} = \eta_1 \mathcal{P}\{x_b \mid H_1\}$, and $\mathcal{P}\{x_a \mid H_0\} = \eta_0 \mathcal{P}\{x_b \mid H_0\}$, and $\eta_1 > \eta_0$. Then it follows that $x_a$ should be ranked higher than $x_b$ on the true ROC curve. If samples are obtained such that the estimated ranking of $x_a$ and $x_b$ are reversed, then the true operating point will move relative to the estimated operating point by $\Delta P_d = \mathcal{P}\{x_a \mid H_1\}(1 - \eta_1)/\eta_1$, $\Delta P_f = \mathcal{P}\{x_a \mid H_0\}(1 - \eta_0)/\eta_0$. Interchanging two operating points does not necessarily result in a change in the ROC, however. An appropriate metric capable of measuring the deviation of the curve must be used. A simple metric is the change in slope of the ROC curve at the point where the two bins are incorporated (the slope of the true ROC curve at a value of $P_f$ is always maximal):

$$\Delta P_d / \Delta P_f = \frac{\mathcal{P}\{x_a \mid H_1\}}{\mathcal{P}\{x_a \mid H_0\}} \frac{(1 - \eta_1)}{(1 - \eta_0)} \frac{\eta_0}{\eta_1} \quad (7)$$

It can be verified that this change reflects a major shift from the convex curve spanned by the true ROC when (i) the bin sort orders are reversed, (ii) there is a significant difference in the true likelihood ratios of the two histogram bins and (iii), there is a large absolute difference in the magnitude of the two class histograms for the respective bins.

## 3 ROCs FROM FINITE DATA

In practice, the class conditional distributions have to be estimated from a finite labeled data set. Formally, we consider the set of all possible class conditional densities as a sample space $\Theta$. Each classification problem is generated by sampling two elements from $\Theta$, yielding the values $\theta_{j|Hi} = p(x = j|H_i), j = 0, 1, \ldots L - 1, i = 0, 1$ where $L = 2^l$ corresponding to the bin probabilities of the class histograms. We assume the histograms for the two class distributions to be independent, i.e. given the class, the features of two different samples will be independent (there may be dependence amongst the features for a given sample).

A finite data set is then drawn independently from each of these class conditional distributions, with $n_i$ samples yielding $k_{j|Hi}$ successes (occurrences) of symbol $x_j$ for class $H_i, i = 0, 1$. We assume the class data is labeled correctly. From this data set we generate class conditional histograms (or density estimates $\hat{\theta}_{j|Hi} = k_{j|Hi}/n_i$), and likelihood ratio estimates

$$\hat{\zeta}_j = \hat{\theta}_{j|H1}/\hat{\theta}_{j|H0} \quad (8)$$

The NP procedure in Section 2 is then performed using these estimates, yielding a set of classifiers and an estimated ROC curve.

There are then two sources of error: (i) errors in the estimates $\hat{P}_d$ and $\hat{P}_f$ for a given classifier $\Gamma$, and (ii), errors due to searching the wrong set of classifiers because the likelihood ranking was incorrect. As an illustration of the problem, consider the simplest case where we wish to separate the two classes in the example above. In this case we obtained 40 samples for each of the two classes from the true distributions, calculated the class-conditional histograms, and performed the Neyman-Pearson procedure using the estimated data. The number of samples for each feature vector were as follows:

| $j$ | 0 | 1 | 2 | 3 |
|---|---|---|---|---|
| $k_{j|H1}$ | 18 | 10 | 5 | 7 |
| $k_{j|H0}$ | 6 | 13 | 12 | 9 |

The true operating curve (TOC) and the estimated performance curve (EPC) that would be achieved using the five classifiers generated by the NP procedure were also calculated. The effect of the two types of errors can clearly be seen in Figure 1.

We now proceed to quantify how errors in the density estimates influence the ROC computation.

### 3.1 HISTOGRAM STATISTICS

It is well known that the bins of each class histogram satisfy the multinomial distribution [7,8]:

$$\mathcal{P}\left\{k_{0|Hi}, \ldots k_{L-1|Hi} \mid \theta_{0|Hi}, \ldots \theta_{L-1|Hi}\right\}$$
$$= \frac{n_i!}{k_{0|Hi}, \ldots k_{L-1|Hi}} \prod_{j=0}^{L-1} (\theta_{j|Hi})^{k_{j|Hi}} \quad (9)$$

where $L = 2^l$. This distribution is a generalization of the binomial distribution; every marginal is also multinomial, and each variable is binomially distributed. Note that the variables are correlated; this correlation becomes small for relatively flat distributions as $l$ increases.

Since the class distributions are independent, we can analyze the statistics of each histogram independently. For notational simplicity, we do not explicitly indicate the class unless necessary.

Our first objective is to be able to bound the error in the estimate of performance for each classifier, when the true values $\theta$ are unknown. We are therefore interested in obtaining the distribution $\mathcal{P}\left\{\theta \mid \hat{\theta}\right\}$, i.e., once we have observed the data, what do we know about the true class distributions based on which samples were generated? To establish valid confidence regions, a prior on $\theta$ has to be assumed[2]. We select the maximum entropy prior, where $\theta$ is uniformly distributed over the probability simplex $\sigma(L) \subseteq [0,1]^L$, and consider a neighborhood $C(\hat{\theta})$ centered on $\hat{\theta}$. Given an observation $\hat{\theta}$, we can then obtain the following Bayesian inversion:

$$\mathcal{P}\left\{\theta \notin C(\hat{\theta}) | \hat{\theta}\right\} = \frac{\int_{[0,1]^L \backslash C(\hat{\theta})} \mathcal{P}\left\{\hat{\theta} \mid \theta\right\} \psi(\theta) \, d\theta}{\int_{[0,1]^L} \mathcal{P}\left\{\hat{\theta} \mid \theta\right\} \psi(\theta) \, d\theta}$$
$$= \frac{\int_{\sigma(L) \backslash C(\hat{\theta})} \prod_{j=0}^{L-1} \theta_j^{\hat{\theta}_j n} \, d\theta}{\int_{\sigma(L)} \prod_{j=0}^{L-1} \theta_j^{\hat{\theta}_j n} \, d\theta} \quad (10)$$

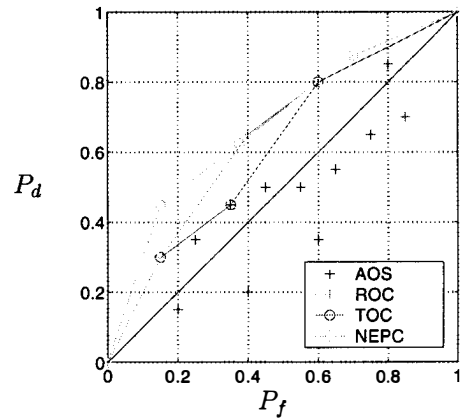

Fig. 1. Set of achievable operating points (AOS), the ROC curve, naive estimated performance curve (NEPC) based on frequency counts, and true operating curve (TOC) achieved by the set of classifiers selected using the Neyman-Pearson procedure on the sample estimated data.

From this integral it can be recognized that the posterior distribution for $\theta|\hat{\theta}$ is a generalized multivariate *beta* density [9]. Note that the density is continuous, conditioned on a discrete variable. The marginals $p(\theta_j|\hat{\theta}_j)$ for a single bin $j$ is given by a $beta(n, k_j)$ distribution where $k_j = n\hat{\theta}_j$, denoted by

$$p(\theta_j|\hat{\theta}_j) = \frac{\theta^{k_j}(1-\theta)^{n-k_j}}{\beta(k_j+1, n-k_j+1)} \quad (11)$$

where $\beta(p, q)$ denotes the normalizing (beta) function.

---

[2] It can be shown that without a prior, no finite amount of data can produce a non-trivial bound on $\theta$ given only $\hat{\theta}$

The mean, mode, and variance are given by

$$E\{\theta_j\} = \mu_j = \frac{k_j + 1}{n + 2} \quad (12)$$

$$\arg\max_\theta p(\theta_j) = \frac{k_j}{n} \quad (13)$$

$$\sigma_j^2 = \frac{(k_j + 1)(n - k_j + 1)}{(n+2)^2(n+3)} \quad (14)$$

When $n \geq 1$ and $n - k_j \geq 1$ the function is unimodal with a bell shape. When $k_j = 0$ or $(k_j = n)$, the distribution is one-sided decaying to the right (left) from 0 (1). Figure 2 shows the exact width $w_{90} \simeq 2\nu$ of the interval starting at the 5% percentile, and ending at the 95% percentile of various *beta* distributions. A Chebychev bound on the tails of the distribution using the variance estimate exist,

$$\mathcal{P}\{|\theta_j - \mu_j| > \nu \mid k_j, n\} \leq \frac{\hat{\theta}_j(1 - \hat{\theta}_j)}{n\nu^2} \leq \frac{1}{4n\nu^2} \quad (15)$$

with a $1/n$ convergence rate (Chernoff bounds giving exponential convergence exist – the Chebychev bound is sufficient for our purposes).

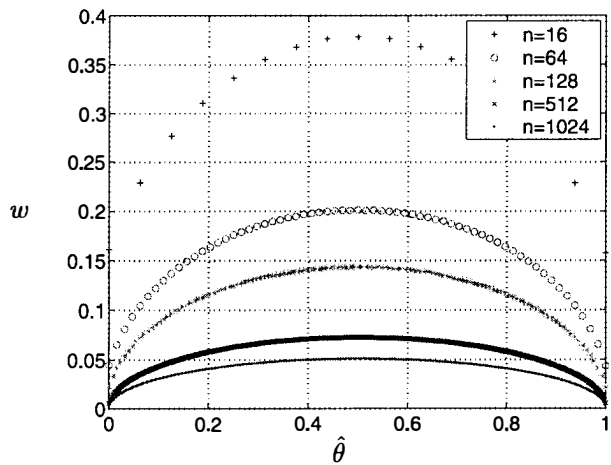

Fig. 2. Plot of the width $w_{90}$ of the $[0.05, 0.95]$ percentile interval of the beta distribution for various values of $\hat{\theta}$ and number of samples $N$.

### 3.2 INFLUENCE OF SAMPLE SIZE ON PERFORMANCE EVALUATION

The results above can be used to quantify how accurately the operating point of a classifier can be estimated. Assume that we are given a particular classifier $\Gamma$, i.e. labels have been assigned to each of the histogram bins. Given *independent* sample data sets from the two classes, from (2) we estimate the detection rate by adding all the histogram entries corresponding to the positive decision region of the classifier:

$$\hat{P}_d = \sum_{j \in \mathcal{L}(\Gamma)_1} \hat{\theta}_{j|H_1} \quad (16)$$

From (9) and (16), the estimated $P_d$ is therefore given by the sum of $\#\mathcal{L}(\Gamma)_1$ of $L$ multinomial variables. Comparison with the formulation of Section 3 will convince the reader that the estimated statistics of $\hat{P}_d$ is the same as that of a histogram with two bins, where the first bin has a count equal to the sum of the counts of the bins in $\mathcal{L}(\Gamma)_1$, and the second bin contains the sum of the bins in $\mathcal{L}(\Gamma)_0$. The estimate $n_1 \hat{P}_d$ is therefore distributed binomially with parameter $\sum_{j \in \mathcal{L}(\Gamma)_1} \theta_j = P_d$. The posterior distribution $\mathcal{P}\{P_d \mid \hat{P}_d\} \simeq beta(n_1, n_1 \hat{P}_d)$. The estimated false alarm rate $\hat{P}_f$ can be analyzed in a similar fashion, yielding $\mathcal{P}\{P_f \mid \hat{P}_f\} \simeq beta(n_0, n_0 \hat{P}_f)$.

The importance of this result is that for a given classifier $\Gamma$, bounding the location of the true operating point $(P_f(\Gamma), P_d(\Gamma))$ around the estimated operating point $(\hat{P}_f(\Gamma), \hat{P}_d(\Gamma))$ can be done to arbitrary confidence levels purely as a function of the number of data samples $n_0$ and $n_1$ used for evaluation. This follows directly from the Chebychev variance bound (15) which does not involve the size of the alphabet (number of features). For example, localization of over 90% probability of the posterior probability occurs in the interval $P_d \pm 2.5\%$ when $n_1 = 1024$. This result is consistent with more general theories of generalization such as PAC-learning theory.

We caution the user that the result in this section is conditioned on independence of the design regions $\mathcal{L}(\Gamma)_j, j = 0, 1$ and the data samples used to evaluate performance. Using the design data inappropriately to generate an estimated performance curve (a curve which we will call NEPC – Naive Estimated Performance Curve) can yield bizarre performance results. We discuss this question further in the next section.

### 3.3 INFLUENCE OF ERROR ON IDENTIFYING THE ROC

The above result that highly accurate performance estimation of a classifier is possible for arbitrarily large feature sets once a certain number of samples has been obtained appears to be counterintuitive. As the number of features increase, the number of feature vectors that can occur grows exponentially. Whenever $l > \log_2(n_i)$ there is not even one sample for each histogram bin in the density estimate for class $H_i$; how can the estimate be accurate? Can we really expect that using a few finite data sets we can keep characterizing the ROC curve for larger and larger feature subsets?

The resolution of this conundrum lies in the fact that while the *average* performance of each possible classifier can in fact be accurately estimated, the NP sorting procedure is affected by the alphabet sizes. If we ignore the NP design procedure and evaluate the perfor-mance $(\hat{P}_f, \hat{P}_d)$ of all $2^{2^{\#Q}}$ classifiers for every subset $Q \subseteq Q^*$ using a few independent data sets of sufficient size $n_0$ and $n_1$, we can then locate the ROC curves for every subset $Q$ with arbitrary probability and accuracy, and find the optimal subset of ROC support classifiers. However, NP design requires the evaluation of the performance of only a subset of the set of classifiers (an exponentially small fraction). By using Neyman-Pearson design, the possibility of evaluating a set of sub-optimal classifiers will increase. We discuss this problem in detail in the next section.

### 3.4 STATISTICS ON THE LIKELIHOOD RATIO RANKING

An error in the set of classifiers produced by the NP design results when the sort is incorrect. To quantify this error requires specifying

$$\mathcal{P}\left\{\zeta_0 \leq \zeta_1 \ldots \leq \zeta_{L-1} \mid \hat{\zeta}_0 \ldots \leq \hat{\zeta}_{L-1}, \hat{\zeta}_0, \ldots \hat{\zeta}_{L-1}\right\} \quad (17)$$

The joint statistics on the likelihoods required for this calculation are intractable. We proceed to bound the error from above using the marginal likelihood in each bin, essentially by pairwise comparison of likelihood estimates. Let $\mathcal{F} = \{\zeta_0 \leq \zeta_1 \ldots \leq \zeta_{L-1}\}$, be an event on the joint space and set the following events on the marginals, corresponding to the value of $\zeta$ falling in a specific interval $V_j = \{\zeta_j \in [\gamma_j, \gamma_{j+1}]\}, \forall j : \gamma_j \leq \gamma_{j+1}$. Then it follows that if $\gamma_j \leq \gamma_{j+1} \forall j$, that

$$\begin{aligned} V &= \cap_{j=0}^{L-1} V_j \subseteq \mathcal{F} \Rightarrow \mathcal{P}\{\mathcal{F}\} \geq \mathcal{P}\{V\} \\ \Rightarrow \mathcal{P}\{\mathcal{F}^c\} &\leq \mathcal{P}\{V^c\} = \mathcal{P}\{(\cap_{j=0}^{L-1} V_j)^c\} = \mathcal{P}\{\cup_{j=0}^{L-1} V_j^c\} \\ \Rightarrow \mathcal{P}\{\mathcal{F}^c\} &\leq \sum_{j=1}^{L} \mathcal{P}\{V_j^c\} \end{aligned} \quad (18)$$

The above expression allows us to bound the error $\epsilon$ on the sort by selecting a set of separating thresholds between the modes of the posterior likelihood distributions, and calculating for each posterior likelihood ratio the probability that it violates its bounding thresholds using the posterior marginal distribution:

$$\mathcal{P}\left\{\epsilon \mid \hat{\zeta}\right\} \leq \min_{\gamma} \sum_{j=0}^{L-1} \mathcal{P}\left\{\zeta_j \notin [\gamma_j, \gamma_{j+1}] \mid \hat{\zeta}_j\right\} \quad (19)$$

where $\gamma_j \leq \gamma_{j+1}, j = 0, 1, \ldots L - 1$.

To calculate the marginal distribution in each bin, assume that a total of $k_{j|H_i}$ successes were obtained in $n_i$ samples for bin $j$ in class $H_i$. By assumption the two bin probabilities $\theta_{j|H1}$ and $\theta_{j|H0}$ are independent, hence the density of the ratio $\zeta_j = \theta_{j|H1}/\theta_{j|H0}$ is given using the beta densities (11) of the bin counts. Abusing notation slightly,

$$p(\zeta_j | \hat{\zeta}_j) = \int_{-\infty}^{\infty} |x| \, p_{\theta_j | \hat{\theta}_j, H_1}(x) p_{\theta_j | \hat{\theta}_j, H_0}(\zeta_j x) \, dx \quad (20)$$

Closed form bounds for the ratio of two *beta* densities can be obtained in certain cases [9](p259), but these are complicated. In practice, the posterior density of the ratio $\zeta$ is more easily calculated numerically using (20).

Examples of the *beta* distributions and the posterior distributions on the likelihood ratios for each of the bins in the example problem are shown in Figure 3. In general, when $n_i \gg 1$ and $n_i - k_{j|Hi} \gg 1$ the density of the beta distribution for bin $j$ for class $H_i$ near zero is negligible. The beta distribution can then be approximated by a Gaussian with a mean $\mu_{j|Hi}$ and variance $\sigma^2_{i|Hj}$. It is further possible to approximate the beta variables' ratio by the ratio of two Gaussian distributions one-sided to zero [10]:

$$p(\zeta|\hat{\zeta}) \simeq \frac{(\sigma_0^2 \mu_1 + \sigma_1^2 \mu_0 \zeta)}{\sqrt{2\pi}(\sigma_0^2 + \sigma_1^2 \zeta^2)^{3/2}} \exp\{-\frac{(\mu_0 - \mu_1 \zeta)}{2(\sigma_0^2 + \sigma_1^2 \zeta^2)}\} \quad (21)$$

(where we neglected the explicit bin index $j$ and used the subscript for the class). When both $k_{j|H0}$ and $k_{j|H1}$ becomes small relative to $n_0$ and $n_1$ respectively, this approximation fails. The ratio distribution becomes wide and flat, and significant skew is introduced by the division operator.

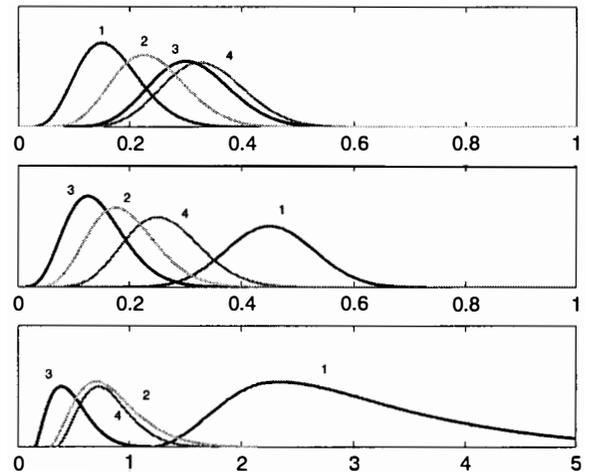

Fig. 3. Top: four beta densities for four bins in the class 0 estimate, with forty samples available for each class. Middle: four beta densities for the bins in the class 1 estimate. Bottom: densities for the likelihood ratios $\zeta$.

## 4 CURSE OF DIMENSIONALITY

The probability of a sort error depends on the exact form of the underlying distributions $\theta_{j|H1}$ and $\theta_{j|H0}$ and general tools for analysis are not available for the discrete case[3]. We restrict our discussion to identifying

---

[3]In the continuous case, some bounds are possible by placing restrictions on the derivatives of the density functions and assuming bounded variations in pointwise estimation error.

conditions under which the correct sort, or a sort with equivalent performance, will be performed with high probability.

From (19) correct sorting of the bins can be guaranteed when the posterior likelihood distributions for the bins do not overlap. This can be guaranteed at some specified level of confidence by ensuring the beta distributions for the two classes in each bin are bounded away from zero, and requiring that the number of samples $n$ be large enough so that via (15) and (21) the posterior distribution spread is substantially less than the distance between the two closest likelihood estimates. In general, using this argument depends on there being a minimum number of samples in every bin of the estimated class distributions where the true probability is non-zero. In the absence of further information about the class distributions, one has to assume that the minimum true bin probability $\theta_j$ will be inversely proportional to $2^l$, the dimension of the space. Hence, guaranteeing the optimal sort may in the worst case require data exponential in the size of the feature set.

In practice, we can argue that two regions of performance occur: a region where performance will degrade slowly with increasing $l$ up to some limit, and a region beyond which the above argument will not apply.

We note that sort errors can be tolerated if the difference of the TOC relative to the ROC is small. When the number of histogram bins with small counts contain only a fraction of the total probability, errors in sorting these will have a limited impact relative to the optimal classifier performance. Neglecting these bins, the remaining beta distributions and their likelihood posterior distributions will be sharply localized. The relative order of bins with large differences in true likelihood can be forced to be correct with high probability, from (19). Bins whose true likelihood values are close may be confused but any resulting shift of the TOC will be small (from (7)). Therefore, in this case either the correct set of classifiers will be found, or classifiers with equivalent performance.

If most of the mass of the distributions remains concentrated in a few bins as the feature dimension increase, the above argument applies and performance should degrade slowly. However, intuitively, when we add irrelevant or weakly correlated features, we expect a randomization effect such that in general most bins will contain only a few samples, and the estimated densities cannot be distinguished from samples from uniform class densities. The posterior distributions for these near-empty bins will be one-sided *beta* distributions overlapping at zero (or infinity). As a result, the probability of all sorts $\alpha$ that differ only due to permutations of these bins will be equally likely. In this case the sort will effectively perform random assignment of labels, and we expect to find classifiers whose true performance approaches the $P_f = P_d$ curve.

In general, therefore, we expect the difficulty of finding the optimal ROC support classifiers to increase with the fraction of the true overall class probability that is found in bins with posterior (beta) distributions near zero. For most distributions when $l \geq \log_2(n)$ most bins will be empty and reflect low confidence estimates. Hence, it is extremely likely that one would obtain a classifier set whose TOC is shifted from the true ROC curve in $P_d$ ($P_f$) corresponding to the density of class 1 (0) in the low-count bins, which can reflect a substantial error.

We now illustrate these results by extending our simple example. We add features $x_2, x_3, x_4 \ldots x_{l-1}$, where for both classes the additional features are uniformly distributed, white variables. The ROC curve for each of the subsets containing $(x_0, x_1, \ldots x_{l-1})$ where $l >= 2$ is therefore equal to the ROC on subset $(x_0, x_1)$, as shown in Fig 1.

For every subset $(x_0, x_1, \ldots x_{l-1})$ we obtained 1024 samples from each of the two classes. A complete NP design was performed using the data set, yielding $2^l$ estimated ROC support classifiers. For each support classifier, we calculated the naive estimated performance curve using this data set (NEPC). In addition, we obtained an additional independent 2048 samples from each class, and calculated the estimated performance of each classifier (denoted EPC for independent estimated performance curve).

The procedure above was repeated 4000 times, for every value of $l$. The results are shown in Figure 4. The first striking result is that as $l$ increases, the EPC and NEPC diverge. Second, the NEPC and EPC curves are well localized at each value of $P_f$ for a fixed $l$. Third, the NEPC increasingly exceeds the true achievable performance (by comparison to Figure 1). Fourth, the EPC approaches the $P_d = P_f$ diagonal as $l$ increases.

Using the result of Section 3.2 we find that the number of data samples $n_0$ and $n_1$ is in principle sufficient to sharply concentrate the posterior distributions of the histogram bins around performance estimates computed with independent sample sets. An ill considered interpretation of the results of Section 3.2 would cause one to also expect the NEPC curves to cluster around the true classifier performance points. However, in evaluating the NEPC, we used one data set to both design, and evaluate the classifiers. The design regions $\mathcal{L}(\Gamma)_j, j = 0, 1$ and the sample distribution are not independent; the NP design procedure finds the bin labeling that correlates the class labels with the sample likelihood ratio; this biases the NEPC (a sample estimate) to exceed the true achievable performance obtained by the classifier. One expects the NEPC to show improved performance as $l$ increases, since the probability that corresponding bins in the two class histograms both contain samples will decrease with $l$, which ultimately results in perfect classification of the sample.

The bias in the NEPC can be bounded in terms of the width of the posterior bin distributions; for example, the expected bias is less than the the sum of the expected width of the beta distributions around the mean. When the density estimates are accurate, the NEPC will be close to the true operating point. This bound is unfortunately only useful when the classifier has few active bins, and the posterior bin distributions are highly localized.

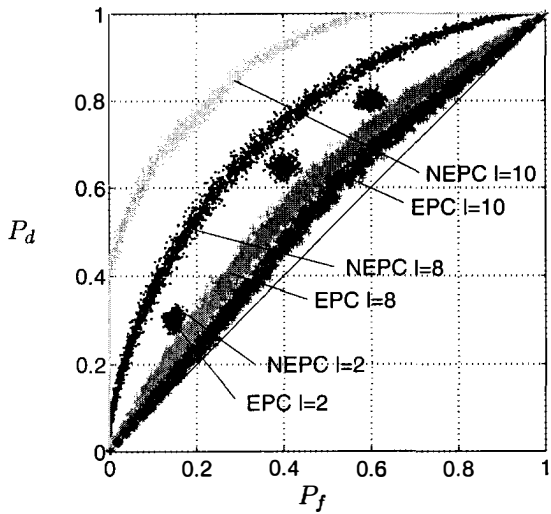

Fig. 4. Estimated operating points (EPC) and Naive Estimated Operating Points (NEPC) obtained as the number of dimensions $l$ increases.

The correct interpretation of the posterior distribution bounds on the estimated performance (15) is that on average, if a *fixed* classifier is evaluated on multiple, independent samples of size $n$, then the average distance from the estimated operating point, to the true operating point, can be made small. Since the true operating point is unique, the estimated performance curves will cluster. The procedure used to evaluate the EPC reflects the correct interpretation of (15) to localize the true operating curve. While we do not discuss this issue further, we note that cross-validation procedures can trade off computational cost to reduce the data requirements for further estimation.

A more subtle point to explain is why the EPC and NEPC curves are well localized over *all 4000 classifier design procedures*. While the operating point for a given classifier can be located accurately, we cannot guarantee (nor in fact are) the same classifiers produced by the NP procedure for each 1024 sample batch. One would therefore expect each NP design to produce radically different EPC and NEPC curves. The explanation for this behavior is that the fraction and location of bins that are well estimated (having collected sufficient samples) remain relatively constant across designs. These bins provide a baseline performance, while the remaining bins are classified essentially randomly and do not contribute in a net way to movement away from the diagonal. We note that other search procedures might not have the same property.

## 5 DATA-ADAPTIVE NP DESIGN

The results in the previous sections can be used to define more statistically sensitive NP-type search procedures. The simplest approach simply randomly classifies histogram bins where the confidence in the class-conditional estimates is too low. The error relative to the true performance curve is limited to the fraction of the overall class probabilities in these bins, and can be estimated. However, this approach does not scale well when the probability mass function does not remain concentrated as the problem dimension increases.

Better results are obtained by using adaptive histograms that merge bins with low confidence estimates, or whose position in the sort is uncertain. By merging bins, small numbers of samples are combined into a bin with a larger effective bin count, concentrating the posterior bin and likelihood distributions.

While we have selected bins on the basis of improving the bound (19) this equation involves pairwise comparisons, and is therefore computationally impractical. More practically, bins have to be selected based only on criteria that can be calculated rapidly for an individual bin. A useful scoring function based on individual bin statistics is

$$t(j) = \frac{||\hat{\theta}_{j|H1} - \hat{\theta}_{j|H0}||}{w_{90}(\hat{\theta}_{j|H1}) + w_{90}(\hat{\theta}_{j|H0})} \qquad (22)$$

where $w_{90}(\hat{\theta}_{j|H1})$ denotes the width of the center 90% interval of the *beta* distribution for the bin $j$. This scoring function measures the theoretical separation achievable for the two classes using only the class conditional counts for the single bin $j$.

Figure 5 shows the result of using this approach where the two lowest scoring bins are merged until some confidence threshold is reached for every bin. For this problem we used $l = 7$ and $n_0 = n_1 = 1024$. The adaptive procedure merged 42 of the histogram bins, significantly improving the sort confidence and the confidence in each of the histogram bins. Due to the increased confidence in the bin counts, the NEPC is much less biased, and the bias can be bounded. When the probability remains concentrated even as the number of features increase, it may be possible to use the NEPC to estimate performance, which could substantially reduce computational requirements over cross-validation approaches. Further exploitation of the statistical results by algorithm design remains an open problem; in particular, using biased performance estimates for searching, as opposed to expensive cross-validation, has not been adequately investigated.

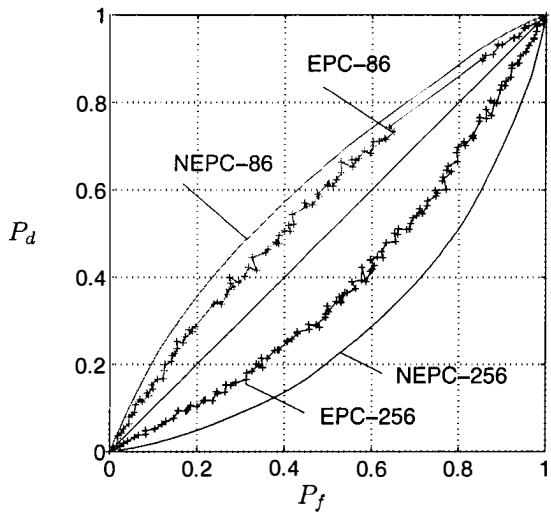

Fig. 5. Estimated operating points (EPC) and Naive Estimated Operating Points (NEPC) obtained with adapting merging of histogram bins. ($P_d$ and $P_f$ are reflected around the diagonal for the 256 bin case to improve clarity).

The above approaches focus on improving the design procedure by improving ROC estimation on a single feature set. It is also possible to use the structure across feature subsets appropriately.

Consider two feature subsets $Q_1$ and $Q_2$. For some sets, an ordering called *uniform preferability* can be established on the ROC curves denoted $\succeq$, where $Q_1 \succeq Q_2$ if $\forall (P_d, P_f) \in \text{ROC}(Q_2) \exists (P'_d, P'_f) \in \text{ROC}(Q_1)$ where $P'_d \geq P_d$ and $P'_f \leq P_f$. Graphically, the ROC curve for feature subset $Q_1$ lies above the ROC curve for feature subset $Q_2$. We use two properties of $\succeq$. First $\succeq$ always defines a filter structure (directed set ordering) on the power set by $Q_1 \supseteq Q_2 \rightarrow Q_1 \succeq Q_2$, since any features not present in the subset will at worst be ignored by the Bayesian fusion procedure (for infinite data or accurately estimated performance curves). Second, given two subsets $Q_1$ and $Q_2$, the ROC of $Q_1 \cup Q_2$ has to be uniformly preferable to the convex hull of the two ROCs, corresponding to sampling classifier structures on $Q_1$ and $Q_2$.

Because of the two properties above, it is possible to define forward feature selection procedures that automatically try to determine the size of the maximal feature subset that can be ranked. By comparing the sets $Q_1$ and $Q_2$ of sizes $l_1$ and $l_2$ respectively, the convex hull of $\text{ROC}(Q_1) \cup \text{ROC}(Q_2)$ can be compared against the ROC estimate obtained on $Q_1 \cup Q_2$ until on average no statistically significant improvement occurs for the larger sets.

Finally, we note that the plethora of existing approaches to future selection (see [3] for an overview) can still be reliably used when the size of the subset under consideration is kept below the threshold required for statistical significance. In particular, backward elimination of irrelevant features from reasonably sized subsets is still indicated [2,11], although in high-dimensional cases, not enough data may be available to consider all features simultaneously.

## 6  CONCLUSIONS

We presented an analysis of the Neyman-Pearson design procedure when applied to finite data sets of discrete variables. The analysis shows that while the performance of a particular classifier assignment can be evaluated independent of the size of the feature set, the probability that the correct sort will be performed by the Neyman-Pearson design procedure decreases as the size of the feature sets increase. Therefore, comparing the ROCs obtained on larger feature subsets with those on smaller feature subsets may not be meaningful; while the operating points for the classifiers spanning the ROC may in each case be almost exact, a sub-optimal set of classifiers will have been found for the larger subset. Therefore, it makes little sense to rank subsets beyond some size, and feature selection procedures should only search a subset of the feature power set.